\documentclass{article}
\PassOptionsToPackage{numbers,sort&compress}{natbib}
\usepackage[preprint]{neurips_2026}
\usepackage[utf8]{inputenc}
\usepackage[T1]{fontenc}
\usepackage{amsmath, amssymb}
\usepackage{microtype}
\usepackage{booktabs}
\usepackage{graphicx}
\usepackage{tikz}
\usetikzlibrary{arrows.meta, positioning, shapes.geometric}
\definecolor{cInk}{HTML}{1F2937}
\definecolor{cAgent}{HTML}{1E3A5F}
\definecolor{cAccent}{HTML}{B85042}
\definecolor{cAmber}{HTML}{BA7517}
\usepackage{hyperref}

\title{Managing Action Preconditions in Neuro-Symbolic RL: Three Placement Strategies for Embodied Agents}
\workshoptitle{Neuro-Symbolic Embodied Intelligence}

\author{%
	Norbert Oswald\\
	Institute of Distributed Intelligent Systems\\
	University of the Bundeswehr Munich\\
	Munich, Germany \\
	\texttt{norbert.oswald@unibw.de} \\
	\And
	Fabian Deuser\\
	Institute of Distributed Intelligent Systems\\
	University of the Bundeswehr Munich\\
	Munich, Germany \\
	\texttt{fabian.deuser@unibw.de} \\	
	\And
	Thomas Bräunl\\
	School of Engineering\\
	The University of Western Australia\\
	Perth, Australia\\
	\texttt{Thomas.Braunl@UWA.edu.au}\\
}

\begin{document}

\maketitle

\begin{abstract}
\noindent
Humans carry behaviour knowledge of how to act in familiar situations into every new task rather than relearning it from scratch. There is no reason a Reinforcement Learning (RL) agent shouldn't do the same: known behaviour patterns need not be learned, only applied. Neuro-symbolic RL bridges prior knowledge and RL by injecting symbolic knowledge alongside a learned policy. The point at which this knowledge is integrated is critical: a poor choice can produce, for instance, hallucinated preconditions, which surface as safety and reliability problems in agents acting in changing environments. We formalise this behavioural knowledge as a precondition Bayesian network (BN) over the agent's \emph{structural
actions} - the actions whose legality depends on preconditions, such as picking up a key, grasping a block, 
toggling a door, or dropping an object. The BN restricts when these actions may fire, and we inject it into the 
RL loop at three placements: (1) a \emph{symbolic verifier}, consulted only at inference, that fires a structural
action once its preconditions hold; (2) a \emph{symbolic enforcer}, active during both training and inference, 
that governs structural-action use throughout learning; and (3) a \emph{symbolic learner}, which folds the 
knowledge into the network and learns the restriction and use of structural actions itself. To test the three variants we run experiments on two benchmarks with opposite regimes: one built on long, ordered
planning chains, the other on continuous manipulation. We compare against strong baselines on solution quality, 
sample efficiency, and traceability. The payoff is substantial. On MiniGrid, all three placements improve the \emph{solution quality} over the PPO+RND baseline, the symbolic enforcer leading at $98.2\%$ against the baseline's $88.8\%$. 
On Fetch, where SAC+HER already solves the task at about $97\%$, solution quality no longer separates the 
placements. The payoff shifts to \emph{sample efficiency}, with symbolic enforcer and learner reaching that ceiling 
about $2\times$ sooner. The advantage also holds beyond the embodied benchmarks: on a non-embodied routing task over a real street network, external placement again leads and generalises to held-out instances. Beyond performance, the symbolic verifier and enforcer keep behaviour checkable: a structural action never fires unless its grounded preconditions hold. 
\end{abstract}

\section{Introduction}
\label{sec:intro}
A reinforcement-learning agent that must discover everything by trial and error learns slowly, and in large or 
sparse-reward environments may not learn at all. Prior knowledge of how to act - which actions are legal in a 
given situation, and what they achieve - lets the agent apply what it already knows instead of rediscovering it, 
so it explores less and learns faster and better, especially in complex environments. This raises three 
questions: what an agent needs to know, how to represent it, and where in the RL loop that knowledge should enter, 
the last of which, our focus, has received far less attention than the first two.

Robotic and interactive agents that act in changing environments must reason over goals and constraints, execute long action sequences, and recover when assumptions fail. In this setting, hallucinated preconditions or
brittle domain models become concrete safety and reliability
problems: an agent that fires a structural action without its preconditions holding may cause irreversible errors, e.g.\ dropped or damaged objects. Neuro-symbolic reinforcement learning~\cite{kautz2022thirdai} offers a principled response to these kinds of problems, injecting behavioural knowledge about action preconditions and effects into an otherwise learned policy. We use the term behavioural knowledge for prior knowledge about how to act in familiar situations: the preconditions and effects of actions and the behaviour patterns that follow from them, independent of whether it is hand-specified or discovered from data. We formalise this behavioural knowledge as a \emph{precondition Bayesian network} (BN) over states, predicates, subgoals, and actions. 
What remains open is the appropriate injection point: whether the agent applies this knowledge at inference, is trained under it, or learns it into its own weights. When the domain model is given, as is often the case in service-robot, warehouse, or assistive settings where task structure is well understood, the design question is where to inject symbolic knowledge, not how to discover it.

This paper studies where in the RL pipeline symbolic structure should enter. Three natural placements suggest themselves: external at inference, external during training, and internal to the network via a supervised head. 
To our knowledge, no prior work has compared them on the same precondition specification. We provide that comparison on two benchmarks chosen to span two paradigms of embodied RL, namely discrete sequential planning (MiniGrid ObstructedMaze~\cite{chevalierboisvert2023minigrid}) and continuous control (FetchPickAndPlace~\cite{plappert2018multigoal,gymnasiumrobotics}). We further test whether the resulting ordering carries beyond embodied simulation to a non-embodied routing task on a real street network. Across all three, we analyse the trade-offs in solution quality, sample efficiency, and traceability, 
and find that where the rule enters shapes the outcome. 

\section{Related Work}
\label{sec:related}
Giving reinforcement-learning agents prior knowledge of action preconditions has been pursued for over two 
decades. We survey this work along two axes: how the knowledge is represented, and where it enters the RL loop.

Knowledge has been cast in a range of representational forms. Sohn et al.~\cite{sohn2018taskgraph} encode subtask precondition dependencies as a graph that the purpose-built policy network reads as input, enabling zero-shot transfer to unseen graphs. Q-Cogni~\cite{qcogni} models the causal precondition structure from data (via NOTEARS) as a Bayesian network that guides a tabular Q-learner. Whatever the form, this knowledge can be written by hand or learned from data: learning the operators' preconditions and effects~\cite{silver2022nsrt}, inventing the predicates themselves~\cite{silver2023predinvent}, learning symbolic abstractions from raw input~\cite{asai2018latplan}, or synthesising a PDDL model with an LLM~\cite{guan2023llmpddl}.

The least invasive placement wraps an already-trained policy and filters its actions at test time. 
\emph{Shielding}~\cite{alshiekh2018shielding} synthesises a shield from a safety specification that wraps the 
policy and vetoes or corrects any unsafe action, and can leave training untouched. There exist various extensions to this approach: Jansen et al.~\cite{jansen2020safe} replace the hard safe/unsafe verdict with a probabilistic shield that blocks an action when its probability of eventually violating the safety property is too high; Carr et al.~\cite{carr2022shield} extend shielding to partially observable settings and integrate it tightly with deep RL to speed up learning; Könighofer et al.~\cite{koenighofer2021games} compute the shield at runtime instead of precomputing it. A second placement lets the symbolic structure shape the learning process while staying outside the policy network. Options~\cite{sutton1999options}, HAM~\cite{parr1997ham}, MAXQ~\cite{dietterich2000maxq}, 
HAC~\cite{levy2019hac}, Feudal Nets~\cite{vezhnevets2017feudal} are methods and frameworks for hierarchical RL. They 
split the agent into two levels: a high-level decision layer that picks which subgoal to pursue next, and a low-level decision layer that learns the primitive actions to reach it. Some methods learn the high-level layer (HAC, Feudal Nets) others hand-specify it (HAM, MAXQ). When the high-level layer is a fixed, hand-written rule set, it acts as a symbolic controller that shapes learning from outside the policy. SDRL~\cite{lyu2019strips} is one such case: a STRIPS/PDDL planner sequences subgoals from action preconditions and effects, and a distinct low-level policy is learned for each subgoal. Reward-level methods instead inject structure through the reward: PDDL-guided RL~\cite{illanes2020pddl} rewards following a symbolic plan, and reward machines~\cite{icarte2018rm} specify the task as a finite-state automaton whose transitions emit reward. A third placement compiles the symbolic knowledge into the network. Differentiable neuro-symbolic methods embed logical reasoning in the forward pass, DeepProbLog~\cite{manhaeve2018deepproblog} being the canonical example, while deep symbolic RL~\cite{garnelo2016dsrl}, neural-symbolic policy learning~\cite{jiang2019nlrl}, and language-grounded RL~\cite{chevalierboisvert2018babyai} fuse symbolic structure with learned representations to varying depths. Kautz's neuro-symbolic taxonomy~\cite{kautz2022thirdai} spans exactly this range, from an external symbolic layer composed with a neural policy to tight in-network coupling.

Prior work almost always commits to a single placement, a shield, a hierarchy, or an in-network module - and 
studies it in isolation. We instead treat inference-time, training-time, and network-level injection as three 
points on one axis, compare them head-to-head on the same specification (§\ref{sec:experiments}), and 
characterise when each wins.

\section{Framework: Symbolic Placement Strategies}
\label{sec:framework}

We formalise the three placements as different points at which a \emph{precondition Bayesian network} (BN) enters the RL loop. Our BN is the deterministic special case of the standard BN formalism: conditionals are boolean values, so evaluating the legality mask reduces to a direct Boolean function evaluation rather than probabilistic belief updating. A probabilistic generalisation, e.g. discovered from data via NOTEARS-style structure learning or neuro-symbolic operator learning (NSRTs) would replace the Boolean mask with a soft mask of probabilities. Our framework carries over unchanged.

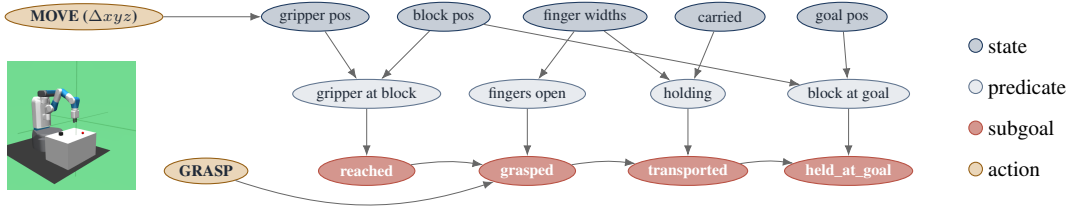
\begin{figure}[h]
	\centering
	\moveright 0.0cm \hbox{
		\begin{minipage}{0.72\textwidth}
			\centering
			\begin{tikzpicture}[scale=0.85, transform shape,
				bnstate/.style={ellipse, draw=cAgent, fill=cAgent!25, text=cInk,
					font=\scriptsize, minimum width=1.4cm, minimum height=0.45cm, inner sep=1pt, line width=0.4pt},
				bnmid/.style={ellipse, draw=cAgent!70, fill=cAgent!10, text=cInk,
					font=\scriptsize, minimum width=1.1cm, minimum height=0.45cm, inner sep=1pt, line width=0.4pt},
				bnsub/.style={ellipse, draw=cAccent, fill=cAccent!60, text=white,
					font=\scriptsize\bfseries, minimum width=1.5cm, minimum height=0.45cm, inner sep=1pt, line width=0.4pt},
				bnact/.style={ellipse, draw=cAmber!80!black, fill=cAmber!30, text=cInk,
					font=\scriptsize\bfseries, minimum width=1.3cm, minimum height=0.4cm, inner sep=1pt, line width=0.4pt},
				bnarr/.style={-{Latex[length=1.4mm]}, line width=0.35pt, draw=black!60}]
				
				\node[bnact]   (move)     at ( -4.4, 5.2) {MOVE ($\Delta xyz$)};
				\node[bnstate] (grippos)  at ( -0.8, 5.2) {gripper pos};
				\node[bnstate] (blockpos) at (  1.2, 5.2) {block pos};
				\node[bnstate] (fingers)  at (  3.4, 5.2) {finger widths};
				\node[bnstate] (carried)  at (  5.5, 5.2) {carried};
				\node[bnstate] (goalpos)  at (  7.4, 5.2) {goal pos};				
				\node[bnmid]   (atblock)  at (  0.0, 4) {gripper at block};
				\node[bnmid]   (fopen)    at (  2.5, 4) {fingers open};
				\node[bnmid]   (holding)  at (  5.0, 4) {holding};
				\node[bnmid]   (atgoal)   at (  7.5, 4) {block at goal};
				\node[bnact]   (grasp)    at ( -2.5, 2.8) {GRASP};
				\node[bnsub]   (reached)  at (  0.0, 2.8) {reached};
				\node[bnsub]   (grasped)  at (  2.5, 2.8) {grasped};
				\node[bnsub]   (transp)   at (  5.0, 2.8) {transported};
				\node[bnsub]   (held)     at (  7.5, 2.8) {held\_at\_goal};
				
				\draw[bnarr] (move) -- (grippos);
				\draw[bnarr] (grippos)  -- (atblock);
				\draw[bnarr] (blockpos) -- (atblock);
				\draw[bnarr] (fingers)  -- (fopen);
				\draw[bnarr] (carried)  -- (holding);
				\draw[bnarr] (fingers)  -- (holding);
				\draw[bnarr] (blockpos) -- (atgoal);
				\draw[bnarr] (goalpos)  -- (atgoal);
				\draw[bnarr] (atblock) -- (reached);
				\draw[bnarr] (fopen)   -- (grasped);
				\draw[bnarr] (holding) -- (transp);
				\draw[bnarr] (atgoal)  -- (held);
				\draw[bnarr] (grasp) to[bend right=18] (grasped);
				\draw[bnarr] (reached) to[bend left=8] (grasped);
				\draw[bnarr] (grasped) to[bend left=8] (transp);
				\draw[bnarr] (transp)  to[bend left=8] (held);
				
				\node[overlay, anchor=south west, inner sep=0pt] at (-5.6, 2.5)
				{\includegraphics[width=2.0cm]{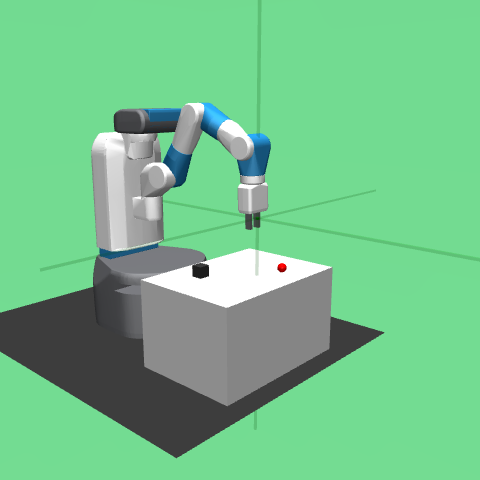}};
				
			\end{tikzpicture}
		\end{minipage}
		\hspace{0.18\textwidth}
		\begin{minipage}{0.2\textwidth}
			\footnotesize
			\raggedright
			\tikz{\node[circle, draw=cAgent, fill=cAgent!25, inner sep=0pt, minimum size=0.18cm, line width=0.4pt] {};}~state\\[0.75em]
			\tikz{\node[circle, draw=cAgent!70, fill=cAgent!10, inner sep=0pt, minimum size=0.18cm, line width=0.4pt] {};}~predicate\\[0.75em]
			\tikz{\node[circle, draw=cAccent, fill=cAccent!60, inner sep=0pt, minimum size=0.18cm, line width=0.4pt] {};}~subgoal\\[0.75em]
			\tikz{\node[circle, draw=cAmber!80!black, fill=cAmber!30, inner sep=0pt, minimum size=0.18cm, line width=0.4pt] {};}~action
		\end{minipage}
	}
	\caption{Precondition BN of FetchPickAndPlace describe active subgoals and when GRASP fires.}
	\label{fig:bn-fetch}
\end{figure}

\subsection{Precondition Bayesian Networks}
\label{sec:precondition-bn}

We consider a standard Markov decision process (MDP) $\mathcal{M} = (\mathcal{S}, \mathcal{A}, \mathcal{P}, \mathcal{R}, \gamma)$ with state space $\mathcal{S}$, action space $\mathcal{A}$, transition and reward function $\mathcal{P}$ and $\mathcal{R}$, and discount factor $\gamma$. 

A \emph{precondition BN} for such an MDP is a directed acyclic graph over four node types and one function:
\begin{itemize}
	\item States $x_1, \ldots, x_n$: raw components of $s \in \mathcal{S}$ or derived quantities cheaply
	computable from $s$, like gripper position (\text{gripper\_pos}), object poses (\text{block\_pos}), or joint velocities.
	
	\item Predicates $\varphi_1, \ldots, \varphi_k$: boolean functions $\varphi_j: \mathcal{S} \to \{0, 1\}$, stacked into the binary predicate vector $\varphi(s)$, e.g. (with threshold $\tau$):
	\begin{align*}
		\varphi_{\text{gripper\_at\_block}}(s) &= \mathbf{1}\bigl[\, \|s_{\text{gripper\_pos}}^{xy} - s_{\text{block\_pos}}^{xy}\|_2 < \tau_{\text{reach}} \,\wedge\, s_{\text{gripper\_pos}}^{z} > s_{\text{block\_pos}}^{z} \,\bigr] \\
		\varphi_{\text{holding}}(s) &= \mathbf{1}\bigl[\, s_{\text{finger\_widths}} < \tau_{\text{closed}}\, \wedge\,
		s_{\text{carried}} \,\bigr]
	\end{align*}
	
	\item Subgoals $\sigma_1, \ldots, \sigma_m$: distinguished predicates that mark task-progress milestones stacked into the binary subgoal vector $\sigma(s)$ (e.g.
	\emph{reached}, \emph{grasped}, \dots). They decompose the task into ordered stages: at each stage the relevant action space can be restricted to actions consistent with the active subgoal g*. 

    \item Actions $a \in \mathcal{A}_{\text{struct}} \cup \mathcal{A}_{\text{nav}}$ split by \emph{effect}:
	\emph{navigation} actions change only the agent's pose, whereas \emph{structural} (manipulative)
	actions change the world state. E.g.:
	\begin{align*}
		\mathcal{A}_{\text{nav}}    = \{\mathrm{left}, \mathrm{right}, \mathrm{forward}\}\\
		\mathcal{A}_{\text{struct}} = \{\mathrm{pickup}, \mathrm{toggle}, \mathrm{drop}\}
	\end{align*}
	
	\item Precondition functions: $\rho_a : \{0, 1\}^{k+m} \to \{0, 1\}$. Each structural action $a \in \mathcal{A}_{\text{struct}}$ is equipped with a precondition function that decides whether $a$ is legal given $\varphi(s)$ and $\sigma(s)$. In practice each $\rho_a$ depends only on $\varphi(s)$ and $\sigma(s)$ that are $a$'s parents in the BN, e.g.
    
	\begin{equation}
		\rho_{\text{GRASP}}(\varphi, \sigma) = \varphi_{\text{fingers\_open}} \wedge \neg
		\varphi_{\text{holding}} \wedge \sigma_{\text{reached}} 
		\label{eq:grasp-pre}
	\end{equation}
\end{itemize}

Figure~\ref{fig:bn-fetch} shows the full Fetch precondition BN. For any state $s$ the BN induces a legality mask vector $\boldsymbol{\rho}(s) = \bigl(\rho_a(\varphi(s), \sigma(s))\bigr)_{a  \in \mathcal{A}_{\text{struct}}} \in {0, 1}^{|\mathcal{A}_{\text{struct}}|}$ collecting the legality bit for each structural action. A structural action $a$ is legal in $s$ iff $\rho_a(\varphi(s), \sigma(s)) = 1$. 

\subsection{Architectures for Knowledge Injection}
\label{sec:arch}
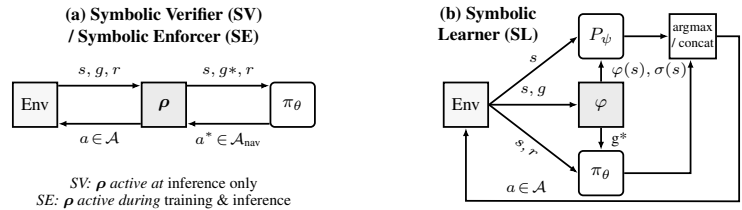
\begin{figure}[h]
	\centering
	\resizebox{0.7\textwidth}{!}{
		\begin{tikzpicture}[
			envbox/.style={draw, thick, minimum height=0.7cm, minimum width=0.7cm, font=\footnotesize, fill=black!3},
			neural/.style={draw, thick, rounded corners=2pt, minimum height=0.7cm, minimum width=0.7cm, font=\footnotesize},
			symbolic/.style={draw, thick, minimum height=0.7cm, minimum width=0.7cm, font=\footnotesize, fill=black!8},
			argbox/.style={draw, thick, minimum height=0.5cm, minimum width=0.7cm, font=\scriptsize},
			label/.style={font=\scriptsize\itshape, align=center},
			title/.style={font=\footnotesize\bfseries},
			fwd/.style={-{Latex[length=1.3mm]}, thick}
			]

			\node[envbox] (b-env) at (0,0) {Env};
			\node[symbolic, right=1.3cm of b-env] (b-pm) {$\boldsymbol{\rho}$};
			\node[neural, right=1.3cm of b-pm] (b-pi) {$\pi_\theta$};
			\draw[fwd] ([yshift=3mm]b-env.east) -- node[above, font=\scriptsize] {$s,g,r$} ([yshift=3mm]b-pm.west);
			\draw[fwd] ([yshift=3mm]b-pm.east) -- node[above, font=\scriptsize] {$s,g*,r$} ([yshift=3mm]b-pi.west);
			\draw[fwd] ([yshift=-3mm]b-pi.west) -- node[below, font=\scriptsize] {$a^*\!\in\!\mathcal{A}_{\text{nav}}$} ([yshift=-3mm]b-pm.east);
			\draw[fwd] ([yshift=-3mm]b-pm.west) -- node[below, font=\scriptsize] {$a\!\in\!\mathcal{A}$} ([yshift=-3mm]b-env.east);
			\node[label, below=0.65cm of b-pm, align=center] {SV: $\boldsymbol{\rho}$ active at \emph{inference only}\\SE: $\boldsymbol{\rho}$ active during \emph{training \& inference}};
			\node[title, above=0.45cm of b-pm, align=center] {(a) Symbolic Verifier (SV)\\/ Symbolic Enforcer (SE)};
			
			\node[envbox, right=2.0cm of b-pi] (c-env) {Env};
			\node[symbolic, right=1.4cm of c-env] (c-pm) {$\varphi$};
			\node[neural, above right=0.35cm and 1.4cm of c-env] (c-p) {$P_\psi$};
			\node[neural, below right=0.35cm and 1.4cm of c-env] (c-pi) {$\pi_\theta$};
			\draw[fwd] (c-env.east) -- node[above, font=\scriptsize, sloped] {$s,g$} (c-pm.west);
			\draw[fwd] (c-env.east) -- node[above, font=\scriptsize] {$s$} (c-p.west);
			\draw[fwd] (c-env.east) -- node[below, font=\scriptsize, sloped] {$s,r$} (c-pi.west);
			\draw[fwd] (c-pm.north) -- node[right, font=\scriptsize, xshift=1pt] {$\varphi(s),\sigma(s)$} (c-p.south);
			\draw[fwd] (c-pm.south) -- node[right, font=\scriptsize, xshift=1pt] {g*} (c-pi.north);
			\node[argbox, right=0.7cm of c-p, minimum height=0.7cm, minimum width=0.7cm, align=center, font=\tiny, inner sep=1pt] (c-comb) {argmax\\/ concat};
			\draw[fwd] (c-p.east) -- (c-comb.west);
			\draw[fwd] (c-pi.east) -| ([xshift=-1pt]c-comb.south);
			\draw[fwd] (c-comb.east) -- ++(0.35,0) -- ++(0,-2.6cm) -- node[above, xshift=-1.2cm, font=\scriptsize] {$a\!\in\!\mathcal{A}$} ([yshift=-1.15cm]c-env.south) -- (c-env.south);
			\node[title, above left=-0.6cm and 0.5cm of c-p, align=center] {(b) Symbolic\\Learner (SL)};
		\end{tikzpicture}
	}
	\caption{Three symbolic placement strategies shown as closed RL loops. $\pi_\theta$ is any RL policy (PPO, SAC, GNN, \dots) wrapped as a black box with subgoal g*; SL additionally attaches a head $P_\psi$.}
	\label{fig:placement-variants}
\end{figure}

To inject the BN into the RL loop, we introduce three architectures shown in Figure~\ref{fig:placement-variants}: a Symbolic Verifier (SV), an inference-only
external wrapper that uses the BN's legality mask vector $\boldsymbol{\rho}(s)$ to override illegal actions; a Symbolic
Enforcer (SE), the same wrapper active also during training to restrict the action space; and a Symbolic
Learner (SL), which network-integrates the BN by feeding predicate vector $\varphi(s)$ and subgoal vector $\sigma(s)$ as input features to 
a supervised head $P_\psi$. Across all variants, $\pi_\theta$  and $P_\psi$ denote neural policies, $a^*$ is the policy's raw action proposal and $a$ the final action returned to the env. All three variants differ only in where the BN is consumed.  

\subsubsection{SV: Symbolic Verifier}
\label{sec:pmw}

The \emph{Symbolic Verifier} (SV) is the simplest and most decoupled of the three variants. Training is
entirely BN-unaware, any standard RL algorithm can serve as $\pi_\theta$. The BN is consulted only at inference: in SV, legality mask vector $\boldsymbol{\rho}(s)$ gates $\pi_\theta$'s outputs, both auto-firing structural actions when preconditions hold and blocking illegal structural firings. At each inference timestep $s_t$, depending on what the environment expects, $\pi_\theta$ produces an action $a_t \in \mathcal{A}$ consisting of a navigation action $a_t^{\text{nav}}$, a structural action $a_t^{\text{struct}}$, or both as a compound tuple. Then, $\boldsymbol{\rho}(s_t)$ modifies $a_t^{\text{struct}}$ as follows before handing over $a_t$ to the environment:
\begin{itemize}
	\item If $\pi_\theta$ proposed a structural action $a_t^{\text{struct}}$ but its precondition function $\rho_{a_t^{\text{struct}}}$ is not satisfied, set $a_t^{\text{struct}}$ to $\emptyset$
	\item If $\pi_\theta$ did not propose a structural action (or an empty one) but there exists a unique legal $a^*\in 
	\mathcal{A}_{\text{struct}}$ whose precondition function is satisfied, replace $a_t^{\text{struct}}$ with $a^*$ 
\end{itemize}

SV borrows the mechanism of shielding~\cite{alshiekh2018shielding}. It is an inference-time gate 
on the policy's actions with different purpose: instead of enforcing safety invariants, it drives task completion, 
firing the structural actions (grasp, drop, toggle) that advance the task. Unlike classical shielding, SV is bidirectional, it can both reject illegal firings and auto-fire unambiguous legal ones.

\subsubsection{SE: Symbolic Enforcer}
\label{sec:ipmw}
Unlike SV, which activates only at inference, the Symbolic Enforcer (SE) consumes the legality mask
$\boldsymbol{\rho}(s)$ throughout training as well. This reduces $\pi_\theta$'s effective action space to
$\mathcal{A}_{\text{nav}}$ and it never has to learn when to fire structural actions: from the very first training step, structural actions are decided by the legality mask $\rho(s)$, not by $\pi_\theta$. Concretely, at each step a structural action fires automatically as soon as its preconditions hold under $\boldsymbol{\rho}(s_t)$, while $\pi_\theta$ supplies only the navigation action $a_t^{\text{nav}}$. On Fetch, where a step is one compound action, $\boldsymbol{\rho}$ sets the structural slot and $\pi_\theta$ sets the navigational one. On MiniGrid, where each step expects a single action, a legal structural action may skip the navigation action; otherwise $\pi_\theta$'s navigation action is submitted. This shrinks $\pi_\theta$'s job to navigation: reach the states where a structural action becomes legal. Because firings happen live during training, the reward for a structural action arrives the moment $\pi_\theta$ reaches the right state, so $\pi_\theta$ learns a short ``reach the state, get the reward'' chain instead of also learning when to fire a structural action.

SE with its two-level hierarchy resembles hierarchical RL (§\ref{sec:related}), where the high-level decision layer is a hand-crafted BN rule set and the low-level layer is $\pi_\theta$. Unlike hierarchies that learn the high level (e.g. 
HAC~\cite{levy2019hac}, Feudal Nets~\cite{vezhnevets2017feudal}), SE's high level has zero learnable parameters. It is cheap and stable, but requires a correct action-precondition specification.

\subsubsection{SL: Symbolic Learner}
\label{sec:ipm}
The Symbolic Learner (SL) integrates symbolic knowledge into the network itself (Figure~\ref{fig:placement-variants}b). Unlike SV, which acts as an external output filter, or SE, which reduces the action space for training, in SL, the predicate and subgoal vectors $\varphi(s), \sigma(s)$ are treated as input features to a learned network head $P_\psi$ which outputs preference values over structural actions. 
Any learned network can serve as $P_\psi$ (such as a small MLP). 

In SL, both networks are trained either separately or jointly, depending on the design decision: a shared or 
separate network. $\pi_\theta$ is trained by the RL reward objective $\mathcal{L}_{\pi_\theta}$ and $P_\psi$ by a supervised cross-entropy loss $\mathcal{L}_{P\psi}$ against the BN. When they share a network, the two losses are combined 
into a single objective, $\mathcal{L}_{\text{SL}} = \mathcal{L}_{\pi_\theta} + \lambda\, \mathcal{L}_{P_\psi}$, and 
optimised jointly.

At inference no external mask is applied. Structural action selection uses argmax over $P_\psi$ and navigation action comes from $\pi_\theta$. But the final action is decided by combining both policies. How they combine depends on what the corresponding environment expects:
\begin{itemize}
\item If nav and structural actions are mutually exclusive (one action per timestep), the two output sets are concatenated and a single argmax over the concatenation selects the action.
\item If nav and structural actions fire concurrently (a compound action tuple per timestep), the compound action is assembled from $\pi_\theta$'s output and the P-head's argmax output.
\end{itemize}

This puts SL on the network-integration side with methods like DeepProbLog~\cite{manhaeve2018deepproblog} and neural-symbolic policy learning~\cite{jiang2019nlrl} rather than shielding: it distils a Boolean rule via gradients instead of enforcing it externally. Because SL learns the legality rule rather than enforcing it, the structural action it selects is not guaranteed to be legal- unlike SV and SE, SL can fire one whose preconditions do not hold.

\section{Experiments}
\label{sec:experiments}
We evaluate the three symbolic placement strategies on two environments drawn from two deliberately different 
paradigms of embodied RL. MiniGrid's ObstructedMaze is a discrete, partially-observed, long-horizon task whose 
difficulty is \emph{strategic}: success demands a specific ordered sequence of structural actions (e.g. toggle box, 
pickup key, drop ball, \dots) under symbolic observations. The MuJoCo-based FetchPickAndPlace is
a continuous-control task whose difficulty is \emph{actuation}: a real-valued end-effector must reach and grasp 
under threshold-defined geometric predicates, with no sequencing to reason about. The two widely-used benchmarks,
spanning discrete planning versus continuous control, thus differ maximally on the two axes that matter here: 
how preconditions are evaluated, and how much precondition structure a task has. For both environments established strong baselines exist. Table~\ref{tab:setup-summary} shows the benchmarks action space as well as chosen baselines. 

\begin{table}[h]
	\centering\footnotesize
	\begin{tabular}{@{}lllcl@{}}
		\toprule
		Benchmark & $\mathcal{A}_{\text{nav}}$ & $|\mathcal{A}_{\text{struct}}|$ & \#Predicates & Baseline \\ 
		\midrule
		MiniGrid     & $\{$left, right, forward$\}$                       & 5:$\{$pickup key/ball, toggle box/door, drop ball$\}$ & $9$ & PPO+RND \\ 		Fetch  & $\Delta xyz$                & 1:$\{$grasp$\}$ & $4$ & SAC$+$HER\\ 
		\bottomrule
	\end{tabular}
	\vspace{1ex}
	\caption{Benchmark configuration summary.}
	\label{tab:setup-summary}
\end{table}

The MiniGrid environment consists of a small two-room gridworld separated by a wall with a locked colour-matched door; the key is hidden inside a closed box and must be revealed. To pick up the blue ball in the second room, the agent must first drop the key (hands must be free). The agent receives no location information, it gets only a $7 \times 7$ egocentric window ahead and acts by moving forward or turning left or right. In total, the observation space is 154-dim, additional structural actions are pickup key or ball, toggle box or door and drop key. The Fetch environment consists of a 7-DOF mobile manipulator on a table with a black cube and a red goal marker. The agent must approach the cube, close the gripper to grasp it, and hold the cube at the goal position, which lies on the table or floats above it. In total, observation space is continuous 25-dim, the navigation action is a 3D Cartesian end-effector velocity and the structural action is grasp.

\subsection{Evaluation}
\label{sec:eval}

We compare the three placements along two axes: solution quality and learning speed. Solution quality measures episode-level task success, scored with each benchmark's native metric. On MiniGrid we report the normalised return: the success rate discounted by $(1 - 0.9\,\bar{t}/t_{\max})$ with mean solved-episode length $\bar{t}$ and horizon $t_{\max}=576$. On Fetch we report the success rate itself: the fraction of evaluation episodes in which the block lies within $0.02$\,m of the goal at episode end. Because both benchmarks resample the task every episode, a new maze layout on MiniGrid, a random block and goal poses on Fetch, these numbers already reflect generalisation across the environment's configuration distribution rather than memorisation of fixed instances. Learning speed measures sample efficiency: the environment steps until a placement variant first reaches $80\%$ of the baseline's final performance, reported as a speed-up ratio. Beyond these two axes we examine traceability qualitatively per domain: BN gating makes every structural firing traceable to the predicate configuration that permitted it, while navigation attribution depends on environment and chosen $\pi_\theta$. 

\subsection{MiniGrid ObstructedMaze}
\label{sec:res-om}
ObstructedMaze (OM-1Dlh) is a sparse-reward, partially-observed maze: the agent sees only a $7\times7$ egocentric window and is rewarded solely on task completion. Under such sparse rewards undirected exploration almost never reaches the goal, so the natural baseline is an exploration-driven method. We use \emph{PPO+RND}: Proximal Policy Optimization~\cite{schulman2017ppo}, an on-policy actor--critic, augmented with Random Network Distillation~\cite{burda2019rnd} for reward shaping. 

\begin{table}[h]
	\centering\footnotesize
	\begin{minipage}[c]{0.3\linewidth}
		\centering
		\includegraphics[width=\linewidth]{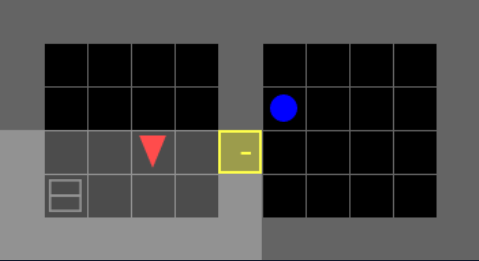}
	\end{minipage}\hfill
	\begin{minipage}[c]{0.7\linewidth}
		\centering
		\begin{tabular}{@{}lccc@{}}
			\toprule
			Method & Env-steps & Success ($\%$) & Norm. return \\
			\midrule
			PPO + RND (baseline) & $10$M & $88.8 \pm 6.1$ & $0.83$ \\
			\hline
			SV: PPO + RND & $10$M & $90.0 \pm 8.4$ & $0.84$ \\
			SE: PPO + RND & $10$M & $98.2 \pm 0.3$ & $0.92$ \\
			SL: PPO + RND & $10$M & $93.8 \pm 7.3$ & $0.88$ \\
			\bottomrule
		\end{tabular}
	\end{minipage}
	\vspace{1ex}

	\caption{ObstructedMaze (left) and results on placement variants with two $\pi_\theta$ compared to baseline.}
	\label{tab:om-results}
\end{table}
In our MiniGrid instantiation, $\pi_\theta$ is the same PPO+RND navigation policy for the baseline and all three 
placements, so the top four rows of Table~\ref{tab:om-results} vary only in where the BN enters. All three placements match or beat the PPO+RND baseline, and where the 
knowledge enters sets the size of the gain. SE is strongest: because it auto-fires the structural actions during 
training, $\pi_\theta$ only has to navigate, and success rises to $98.2\%$ against the baseline's $88.8\%$. SL, 
which must learn the legality rule rather than have it supplied, lands just below at $93.8\%$. SV, correcting the 
already-trained baseline at inference, adds a smaller $+1.2$ pp ($90.0\%$): cheap and never harmful, but unable to
reach states the baseline never learned. 

Traceability comes first from the injected knowledge, not the network: because the BN gates the structural actions, every firing traces to the rule that allowed it independent of the policy used. This is a safety and reliability property, not just a convenience: under SV and SE a structural action is gated on grounded predicates and never fires unless they hold, so the policy cannot act on a hallucinated or ungrounded precondition, and this record makes it a checkable guarantee rather than an assumption. SL relaxes this guarantee, because it never consults the BN at inference but learns to reproduce the rule inside its network, and that learned copy can drift from the true precondition. 

\subsection{FetchPickAndPlace}
\label{sec:res-fetch}
FetchPickAndPlace-v4 is a sparse-reward, goal-conditioned continuous-control task with a
$7$-DoF arm. Where MiniGrid demands discrete planning, the difficulty here is continuous
actuation under a reward that stays silent until success, so the natural baseline is an
off-policy learner with hindsight relabelling: SAC~\cite{sac} with HER~\cite{her}. In our Fetch instantiation all three variants use the SAC+HER policy as $\pi_\theta$ over the continuous $\Delta xyz$ motion and differ only in where the grasp decision enters. SL adds an MLP as $P_\psi$ whose grasp/no-grasp decision concatenates with $\pi_\theta$'s motion into a compound action for the environment. SE instead lets $\rho$ set the grasp slot at every step while SV leaves $\pi_\theta$ untouched, overriding the grasp bit through the legality mask at inference only.
\vspace{-2ex}

\begin{figure}[h]
	\centering
\begin{minipage}[c]{0.45\textwidth}
	\centering	\footnotesize
	\hspace{-6ex}\includegraphics[width=0.91\linewidth]{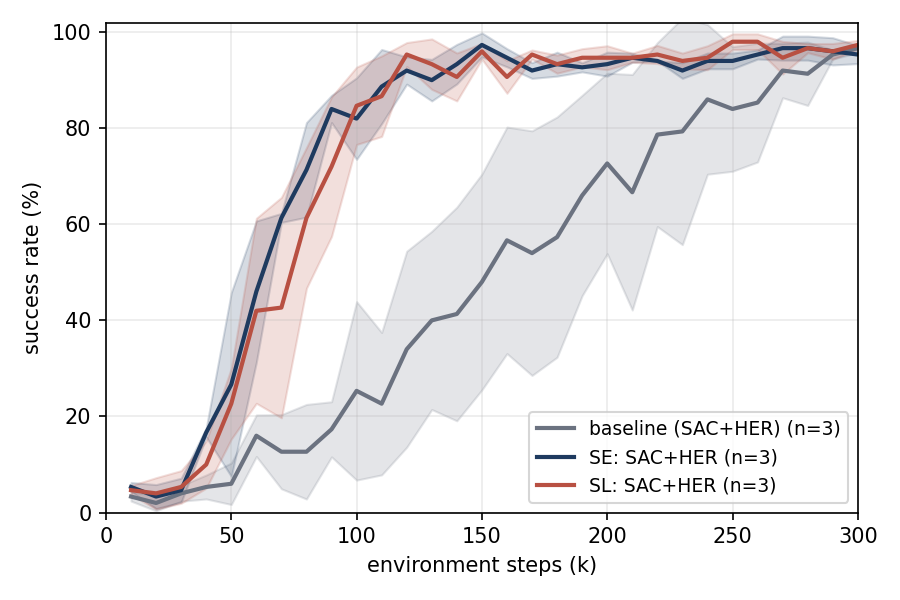}\\[0pt]
	\vspace{-2ex}
	\label{fig:fet-results}
	{\footnotesize (a)}
\end{minipage}\hfill
\begin{minipage}[c]{0.5\textwidth}
		\centering
	\footnotesize
	\setlength{\tabcolsep}{3pt}   
	\hspace*{-8ex}\begin{tabular}{@{}p{1.2cm}p{1.2cm}p{1.2cm}p{3.8cm}@{}}
		\toprule
		Benchmark & Paradigm & Policy $\pi_\theta$ & Payoff of correct placement \\
		\midrule
		MiniGrid & discrete sequential planning & PPO+RND &
		\textbf{Solution quality}: SE reaches $98.2\%$ vs.\ the $88.8\%$ baseline; all placements $\ge$ baseline. \\[2pt]
		Fetch & continuous control & SAC+HER &
		\textbf{Sample efficiency}: $\sim97\%$ ceiling, SE/SL reached $\sim2\times$ sooner\\ 
		Astoria & real-world graph & GNN+HER &
		\textbf{Generalisation}: SV/SE route with held-out endpoints near-optimally\\
		\bottomrule
	\end{tabular}
	\vspace{0.5ex}
	{\footnotesize (b)}	
	\label{tab:summary}
\end{minipage}
\vspace{-1ex}
	\caption{Fetch sample efficiency (a) and cross-benchmark summary (b)}
	\label{fig:fetch}
\end{figure}
			
Unlike MiniGrid, SAC+HER already \emph{solves} this task with about $97\%$ success given enough interactions. Solution quality therefore does not separate the placements since all reach the same success rate. \emph{Sample efficiency} is thus the informative criterion here: how quickly each variant gets there. Figure~\ref{fig:fetch}a shows the learning curves. Lines are the mean over three seeds and shaded bands the standard deviation.
SE and SL reach the baseline's converged success about $2\times$ sooner (roughly at $100$k vs.\ $250$k env-steps) and far more reliably: their bands are tight, while the baseline's band is wide. 
The reason is direct: SE auto-fires the grasp during training and SL supervises it, so $\pi_\theta$ never has to \emph{discover} the grasp, it only has to learn to reach the block. SV, being inference-only, cannot speed up learning, since it wraps an already trained $\pi_\theta$. Traceability is narrower than on MiniGrid, since only one structural action exists. Under SV and SE the gripper closes only when the
grasp preconditions are grounded and hold, so the arm never grasps at nothing.

\subsection{Experiments beyond embodied tasks}
\label{sec:res-nyc}
\begin{figure}[h]
	\footnotesize
	\vspace{0ex}
	\begin{minipage}[c]{0.22\linewidth}  
		\vspace{5ex}
		\includegraphics[width=1.1\linewidth]{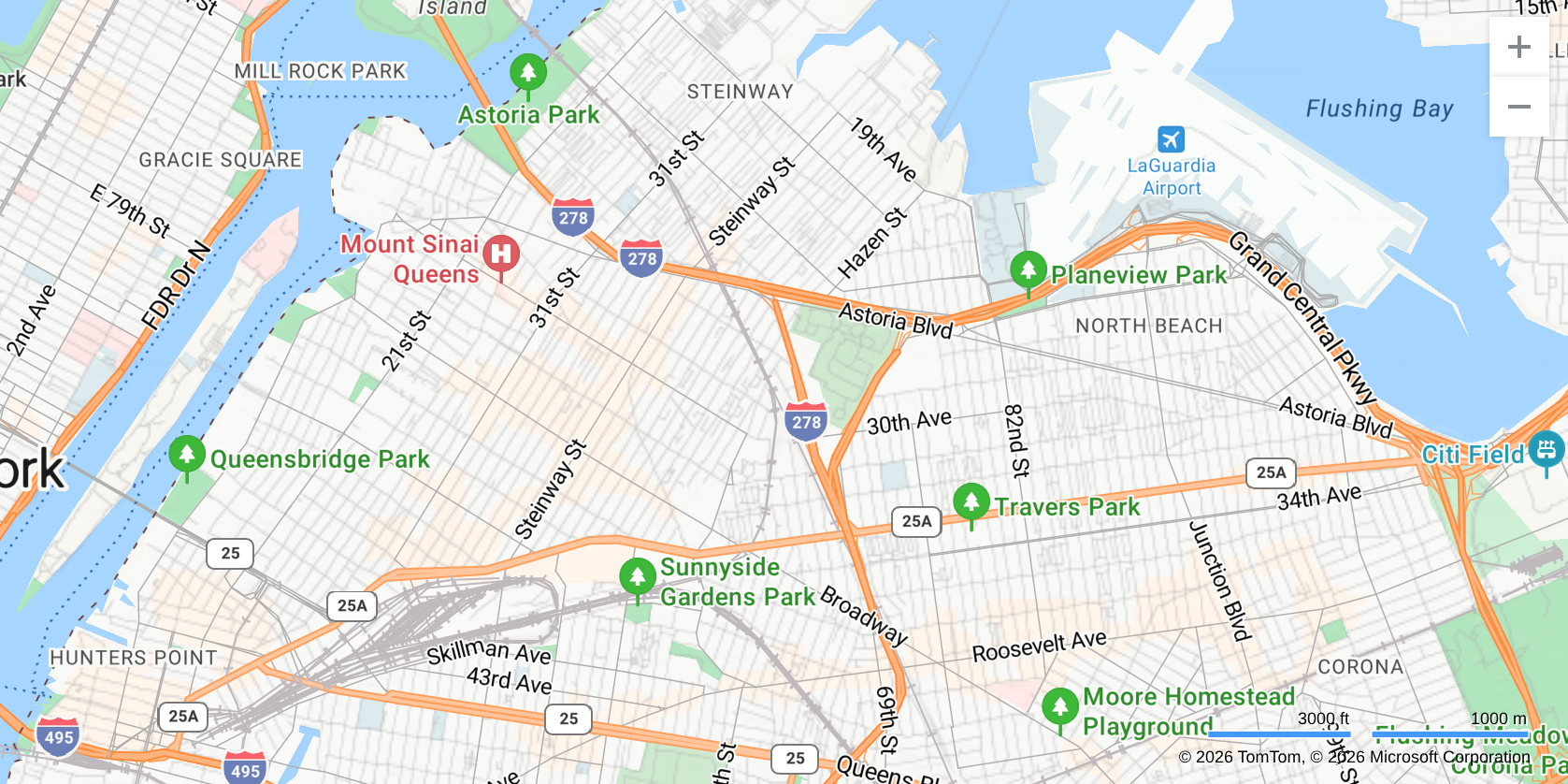}\\
		\vspace{1ex}

		{\footnotesize (a) Astoria street network}
	\end{minipage}
	\hspace{3ex}
	\begin{minipage}[l]{0.50\linewidth}  
		\setlength{\tabcolsep}{3pt}   
		\vspace{5ex}
		\begin{tabular}{@{}lccccc@{}}
		\toprule
		Method & Steps & Success (\%) & Dijkstra & Path ratio & SPL \\
		\midrule
		SV: GNN  & $1$M & $70.8 \pm 2.2$ & $76.4$ & $1.03$ & $0.68$ \\
		SE: GNN  & $1$M & $68.4 \pm 4.4$ & $75.9$ & $1.04$ & $0.66$ \\
		SL: GNN  & $6$M & $64.0 \pm 2.1$ & $40.6$ & $1.11$ & $0.58$ \\
		\bottomrule
		\end{tabular}
		\vspace{4ex}
		
		{\footnotesize (b) Results on \emph{unseen} endpoints never used as training goals.}
	\end{minipage}
	\hspace{-2ex}
	\begin{minipage}[c]{0.25\linewidth}  
		\centering
		\resizebox{\linewidth}{!}{
		\begin{tikzpicture}[
		bnstate/.style={ellipse, draw=cAgent, fill=cAgent!20, text=cInk, font=\scriptsize, minimum width=1.2cm, minimum height=0.4cm, inner sep=1pt, line width=0.4pt},
		bnpred/.style={ellipse, draw=cAgent!70, fill=cAgent!8, text=cInk, font=\scriptsize, minimum width=1.3cm, minimum height=0.4cm, inner sep=1pt, line width=0.4pt},
		bnsub/.style={ellipse, draw=cAccent, fill=cAccent!55, text=white, font=\scriptsize\bfseries, minimum width=1.2cm, minimum height=0.4cm, inner sep=1pt, line width=0.4pt},
		bnact/.style={ellipse, draw=cAmber!80!black, fill=cAmber!30, text=cInk, font=\scriptsize, minimum width=1.0cm, minimum height=0.4cm, inner sep=1pt, line width=0.4pt},
		bnarr/.style={-{Latex[length=1.2mm]}, line width=0.35pt, draw=black!55}]
		\node[bnact]   (move)    at ( 2.5, 4.3) {Move};
		\node[bnstate] (dest)    at (-1.4, 3.4) {Dest};
		\node[bnstate] (taxipos) at ( 1.0, 3.4) {Taxi pos};
		\node[bnstate] (paxpos)  at ( 3.4, 3.4) {Pax pos};
		\node[bnpred]  (atdest)  at (-0.4, 2.2) {Taxi at dest};
		\node[bnpred]  (atpax)   at ( 2.4, 2.2) {Taxi at pax};
		\node[bnsub]   (deliver) at (-0.4, 1.0) {Deliver};
		\node[bnsub]   (paxin)   at ( 2.4, 1.0) {Pax in taxi};
		\node[bnact]   (dropoff) at (-0.4, 0.0) {Dropoff};
		\node[bnact]   (pickup)  at ( 2.4, 0.0) {Pickup};
		\draw[bnarr] (move)    -- (taxipos);
		\draw[bnarr] (dest)    -- (atdest);
		\draw[bnarr] (taxipos) -- (atdest);
		\draw[bnarr] (taxipos) -- (atpax);
		\draw[bnarr] (paxpos)  -- (atpax);
		\draw[bnarr] (atdest)  -- (deliver);
		\draw[bnarr] (atpax)   -- (paxin);
		\draw[bnarr] (pickup)  to[bend right=30] (paxin);
		\draw[bnarr] (paxin)   to[bend left=15] (deliver);
		\draw[bnarr] (dropoff) -- (deliver);
		\end{tikzpicture}}\\
		{\footnotesize (c) Precondition BN}
	\end{minipage}\hfill      
\caption{Astoria street network, graph, and precondition BN~\cite{qcogni}.}
\label{fig:nyc-setup}
\end{figure}
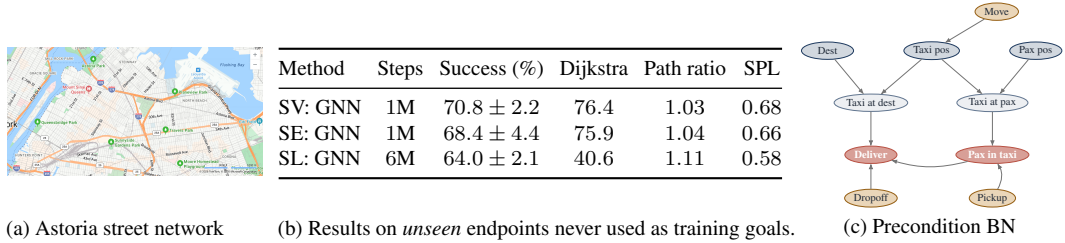

Both environments so far are embodied. To test the three placements beyond embodiment, we apply them to a non-embodied domain built on real-world data: routing over the real Astoria street network of New York City, an OpenStreetMap graph~\cite{osm} of 741 nodes and 1885 directed edges derived from taxi-trip data~\cite{nyctlc}. Each episode starts a taxi at a random node; it must drive to a passenger, pick them up, and deliver them to the 
goal. Formally, a state is the pair (current node, goal), and each step moves to a neighbouring node or fires pickup or dropoff under a sparse reward. The injected behavioural structure is the precondition BN of Cunha et al.~\cite{qcogni} (Figure~\ref{fig:nyc-setup}c). 

All three variants use a GNN+HER policy as $\pi_\theta$ over the street graph. Unlike the embodied tasks, this experiment has no knowledge-free baseline, since we use the policy for two-leg navigation only - from the taxi's node to the passenger, then on to the goal. So pick-up and drop-off must be supplied by one of the placement variants.
That makes the placements a prerequisite here rather than merely an improvement. On the embodied tasks placements lift a policy that already works, whereas on the routing task no single taxi trip completes without them. 
The comparison is therefore between the three placements themselves: which one generalises best. 

We train on an 80/20 trip split and evaluate on 5000 held-out (source, goal) pairs whose endpoints were never training goals, and report success rate, match of agent's route to Dijkstra's (path ratio), as well as success-weighted path length (SPL~\cite{anderson2018spl}). On these held-out trips, SV and SE route near-optimally with path ratio $1.03$ and $1.04$, and with SPL $0.70$ and $0.67$ (Figure \ref{fig:nyc-setup}b). SL routes more coarsely with lowest success rate despite $6\times$ the training budget. Routing these unseen endpoints near-optimally, against Dijkstra at real city scale, is genuine generalisation. 

\section{Discussion and Conclusion}
\label{sec:discussion}
Across all three investigated domains one principle organises our results: when an agent already holds correct behavioural knowledge, e.g. a precondition Bayesian network, the most effective thing to do with it is to apply it as the agent learns, not to spend the network's capacity re-deriving it. All three symbolic placement strategies use the same knowledge, they differ only in where knowledge enters, and across every learner we tried, that choice, not the algorithm or architecture beneath it, drove the results. The general benefits are twofold. First, the agent learns faster and, where the baseline leaves room, better, because it no longer has to discover structure it was handed. Second, only the external placements SV and SE keep behaviour checkable, because a structural action fires only when its preconditions are grounded and true.

Every placement we tried at least matched its baseline but they differed in the kind and size of the gain (Figure~\ref{fig:fetch}b). 
When the baseline already solves the task, as SAC+HER does on Fetch, no placement improves solution quality, all reach the same ceiling. In this case, the gain is sample efficiency instead: SE and SL reach that ceiling about 
$2\times$ sooner, while SV, applied only at inference, merely matches it. When the baselines' success rate is not perfect as on MiniGrid, the placements gain is solution quality, with SE being highest and all are above the baseline. 
On the complex routing task with held-out trips, the gain is generalisation. SV and SE route unseen endpoints near-optimally, whereas SL falls behind, since it must learn the legality rule $\rho$ through weight updates rather than have it supplied.
 
Embodied agents fail not only by acting sub-optimally but also by acting on knowledge that is wrong: a hallucinated precondition, a brittle domain model, an ungrounded affordance, believing something is graspable, reachable, or unlocked when it is not. Reliability means never firing a structural action whose precondition does not hold. Given a valid BN and correctly grounded predicates, SV and SE make this a guarantee for actions the BN covers. SV and SE gate every structural action on those predicates, so an illegal firing cannot occur, and each firing traces back to the exact condition that permitted it. SL offers no such guarantee. Its rule lives in learned weights that only approximate the true precondition and can drift from it.

Besides the mentioned advantages, two limitations are central. First, the framework needs a precondition BN. We 
write ours by hand, but that is a convenience, not a requirement since such structure 
can equally be elicited from an LLM, extracted from datasets, or learned from demonstrations. What the framework 
does require is that the BN be sound. That means a legal structural action always produces its modelled effect. Second, and more practically, structure alone does not, on its own, make complex RL tasks easy. They remain hard RL 
problems, with sparse rewards, long horizons, and large state spaces, and getting any variant to learn still 
required the usual apparatus: finding the right policy and parameters, reward shaping and careful exploration handling. The BN reduces how much of this is needed but it does not remove it.

A further direction is to scale the precondition structure itself. Our BNs are small and flat, which suits a 
handful of predicates, subgoals, and actions. With many more, two pressures grow: on the one hand the cost of injecting preconditions into SL, and on the other hand choosing among many simultaneously-legal structural actions. A hierarchical precondition BN is a natural way to absorb both, mirroring how hierarchical RL decomposes long-horizon control. A potential next move is to take all three placements into longer-horizon embodied tasks such as humanoid loco-manipulation (walk-to, reach-to, grasp-to), and ask which placement does best, and at what cost.

\bibliographystyle{plainnat}
\bibliography{map}

\end{document}